# Learning Selectively Conditioned Forest Structures with Applications to DBNs and Classification

**Brian D. Ziebart**
Machine Learning Department
Carnegie Mellon University
Pittsburgh, PA 15213
bziebart@cs.cmu.edu

**Anind K. Dey**
Human-Computer Interaction Institute
Carnegie Mellon University
Pittsburgh, PA 15213
anind@cs.cmu.edu

**J. Andrew Bagnell**
Robotics Institute
Carnegie Mellon University
Pittsburgh, PA 15213
dbagnell@ri.cmu.edu

## Abstract

Dealing with uncertainty in Bayesian Network structures using maximum a posteriori (MAP) estimation or Bayesian Model Averaging (BMA) is often intractable due to the superexponential number of possible directed, acyclic graphs. When the prior is decomposable, two classes of graphs where efficient learning can take place are tree-structures, and fixed-orderings with limited in-degree. We show how MAP estimates and BMA for selectively conditioned forests (SCF), a combination of these two classes, can be computed efficiently for ordered sets of variables. We apply SCFs to temporal data to learn Dynamic Bayesian Networks having an intra-timestep forest and inter-timestep limited in-degree structure, improving model accuracy over DBNs without the combination of structures. We also apply SCFs to Bayes Net classification to learn selective forest-augmented Naïve Bayes classifiers. We argue that the built-in feature selection of selective augmented Bayes classifiers makes them preferable to similar non-selective classifiers based on empirical evidence.

## 1 Introduction

Bayesian Networks have proven to be useful for modeling relationships between variables. However, optimally and efficiently learning from data which relationships are important for building accurate models (i.e., structure learning) is difficult due to the superexponential number of possible graphs and the requirement of acyclicity. One common approach to structure learning is to find the maximum a posteriori (MAP) estimate for the graph structure of the Bayesian Network. A second approach is to probabilistically aver-age over all possible structures using Bayesian Model Averaging (BMA). In this paper, we provide novel methods for obtaining the MAP structure estimate and BMA for the class of structures with ordered sets of fully observed, discretized variables, where within each set the variables are connected at most by a tree graph (Figure 1a), and each variable can have a limited number of parents from the previous set(s) in the ordering (1b). We call this class of structures *selectively conditioned forests* (SCFs) (1c). We provide an algorithm capable of finding the MAP SCF structure in time that is polynomial in the number of variables in each set, $|\mathcal{V}|$, and the amount of training data, $|D|$, but exponential in $k$, a limit on the number of inter-set parents for each node. The overall runtime of our algorithm is $O(|\mathcal{V}|^{k+2}|D|)$. Additionally, we provide a method for BMA of SCF structures with run time $O(|\mathcal{V}|^{k+2}|D| + |\mathcal{V}|^3)$.

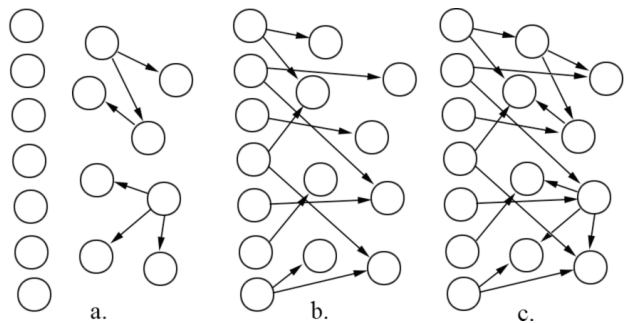

Figure 1: A forest structure (a), a selectively conditioned structure (b), and their combination, a selectively conditioned forest structure (c).

We were influenced to investigate selectively conditioned forests by our work with temporal data modeling problems. When gathering temporal data it is often difficult to observe all of the necessary variables at a high enough temporal granularity to obtain a true causal model of underlying physical phenomena. Even if it were possible, different observation granu-larities between features would result in a very high order model. Instead, some pairs of variables appear to have instantaneous correlation within a timestep. The SCF of a Dynamic Bayesian Network (DBN) contains both sets of inter-timestep and intra-timestep dependencies. The increase in time complexity of finding intra-timestep dependencies in addition to inter-timestep dependencies is only linear in the number of variables when learning the MAP structure. We show that for real datasets, adding intra-timestep structure improves model accuracy more than additional inter-timestep parents even when the intra-timestep variables are unobserved during inference.

As a second application of our method, we generalize Friedman et al. (1997)'s tree-augmented Naïve Bayes classifier by employing MAP SCF to learn selective forest-augmented Naïve Bayes (SFAN) with no increase in asymptotic time complexity. These generalizations to TAN can reduce model parameterization, improving classification accuracy by removing unnecessary dependencies, such as noisy features. While selective Naïve Bayes classifiers have been of interest (Langley and Sage 1994), prior to this work SFAN structures were obtained approximately using heuristics rather than through exact MAP estimation.

The remainder of this paper is organized as follows. In section 2 we review structure learning and Bayesian Model Averaging, and relate our work to similar approaches for dealing with structure uncertainty in DBNs and augmented Naïve Bayes classifiers. We provide the algorithm for efficiently learning the MAP estimate and Bayesian Model Averaging for the SCF class of structures in sections 3 and 4. In sections 5 and 6 we apply MAP SCF to temporal data and classification problems to learn SCF DBNs and selective forest-augmented Naïve Bayes (SFAN) classifiers, and evaluate the resulting models on real datasets to illustrate the benefits of intra-temporal dependencies for temporal modeling and selectivity for classification. Finally, in section 7 we provide conclusions and future extensions for this work.

## 2    Background and Related Work

Bayesian Networks allow for a compact representation of probability distributions for sets of variables. The joint probability of a dataset, $\mathbf{X} = \{\mathbf{X_1}, \mathbf{X_2}, ..., \mathbf{X_n}\}$, factors according to a directed acyclic graph, $G = (\mathcal{V}, \mathcal{E})$, as follows:

$$P(\mathbf{X_1}, ..., \mathbf{X_n}|G, \Theta) \equiv \prod_{i=1}^{|\mathcal{V}|} p(\mathbf{X}_i|\mathbf{X}_{\pi_i}, \theta_i)$$

where the parents of $X_i$, denoted $\pi_i$ are determined by G and have values $\mathbf{X}_{\pi_i}$ (and $X_i$ has values $\mathbf{X}_i$). The conditional probability for $X_i$ given its parents, $\theta_i$, is determined by the set of parameters $\mathbf{\Theta}$.

In many machine learning settings G is unknown and must be learned from data. A common approach to structure learning is to employ a prior on graph structures and parameters, $P(G, \Theta)$, that limits complexity and then use the maximum a posteriori (MAP) estimate for the graph structure.

$$\begin{aligned}
G^*_{MAP} &= \operatorname*{argmax}_G \int_\Theta P(G, \Theta|\mathbf{X}) \, d\Theta \\
&= \operatorname*{argmax}_G \int_\Theta P(\mathbf{X}|G, \Theta) P(G, \Theta) \, d\Theta \\
&= \operatorname*{argmax}_G \prod_i P(\pi_i) \int_{\theta_i} P(\mathbf{X}_i|\mathbf{X}_{\pi_i}, \theta_i) P(\theta_i|\pi_i) \, d\theta_i \\
&= \operatorname*{argmax}_G e^{\sum_i LS(\mathbf{X}_i|\mathbf{X}_{\pi_i}, P_{\theta_i}) + \log P(\pi_i)} \\
&= \operatorname*{argmax}_G \sum_i LS(\mathbf{X}_i|\mathbf{X}_{\pi_i}, P_{\theta_i}) + \log P(\pi_i) \qquad (1)
\end{aligned}$$

$$LS(\mathbf{X}_i|\mathbf{X}_{\pi_i}, P_{\theta_i}) \equiv \log \int_{\theta_i} P(\mathbf{X}_i|\mathbf{X}_{\pi_i}, \theta_i) P(\theta_i|\pi_i) \, d\theta_i \qquad (2)$$

As shown in Equation 1, the structure and parameters priors are usually chosen so that the entire function decomposes into the sum of local score function (LS) evaluations (Equation 2) and local priors, $P_{\theta_i}, P(\pi_i)$, on parameters and structure. Common priors for Bayesian Networks with multinomially distributed variables are the Bayesian Dirichlet (BD), the Bayesian Dirichlet Equivalent (BDe) (Heckerman et al. 1995), and the uniform Bayesian Dirichlet Equivalent (BDeu) (Buntine 1991).

For the general class of directed acyclic graphs, finding the MAP estimate is NP-hard (Chickering 1995) even with decomposable MAP functions. Two restricted classes of structures where the MAP estimate can be found in polynomial time are tree-structures, by reduction to a maximum spanning tree solution with edges weighted by mutual information (Chow and Liu 1968), and fixed-orderings with limited in-degree, by combinatorial search (Buntine 1991). In the latter class of structures, each variable is restricted to have at most k parents occurring previously to itself in a provided ordering of all variables. For less restrictive classes of graphs, heuristic-based edge operations (Heckerman et al. 1995) are employed to search for good structures in polynomial time. However, there are no guarantees that these searches will find the MAP estimate. In contrast, we learn the exact MAP estimate of the combination of tree-structures and fixed-orderings with limited in-degree in polynomial time.

In Bayesian Model Averaging (BMA), inquiries are probabilistically averaged over all possible graph structures and parameters. For example, the probability of new test data, $\tilde{\mathbf{X}}$, can be averaged over all possible models weighted by training data $\mathbf{X}$.

$$
\begin{aligned}
P(\tilde{\mathbf{X}}|\mathbf{X}) &= \sum_G \int_\Theta P(\tilde{\mathbf{X}}|G,\Theta)P(G,\Theta|\mathbf{X})\,d\Theta \\
&= \sum_G \prod_i \int_{\theta_i} P(\tilde{\mathbf{X}}_i|\tilde{\mathbf{X}}_{\pi_i},\theta_i)P(\pi_i,\theta_i|\mathbf{X})\,d\theta_i \\
&= \sum_G \prod_i e^{LS(\tilde{\mathbf{X}}_i|\tilde{\mathbf{X}}_{\pi_i},P_{\theta_i}|\mathbf{X})+\log P(\pi_i)}
\end{aligned}
\tag{3}
$$

For general Bayesian Networks, the superexponential number of possible structures to consider makes Bayesian Model Averaging intractable. However, efficient BMA can be employed on structures with fixed-orderings and limited in-degree (Friedman and Koller 2000), and tree-structures (Meila and Jaakkola 2000, Cerquides and de Mántaras 2003) where the sufficient statistics of $P(G,\Theta|\mathbf{X})$ can be efficiently computed by using a decomposable conjugate prior. For Bayesian Model Averaging of tree-structures, efficient computation relies on a combinatoric result of graph theory, the Matrix Tree Theorem (West 2000). We provide a method to perform Bayesian Model Averaging over the class of SCFs using a generalization of this theorem.

Temporal datasets represent an important domain for structure learning. Dynamic Bayesian Networks (DBNs) are an extension of Bayesian Networks to temporal data of arbitrary length. A DBN consists of a set of variables repeated over many consecutive timesteps. Edges in DBNs are generally restricted from being directed backwards in time. The graph structure for two-slice DBNs consists of a structure for the first timestep, $G_0$, and a structure repeated for every consecutive timestep, $G_+$.

In two-slice DBNs, structure learning has been applied to learn the initial timestep structure, $G_0$, and the repeated structure, $G_+$ (Friedman et al. 1998). The conditional Chow-Liu algorithm (Kirshner et al. 2004) obtains the MAP estimate of the class of structures where each variable can have one parent from either the same or the previous timestep in addition to a fixed set of condition variables. The K2 algorithm (Cooper and Herskovits 1992) is also employed to learn the structure for DBNs with a fixed number of parents from the previous timestep. Conditional Chow-Liu tree structures and limited inter-timestep parents are both subclasses of SCFs, which allow for the combination of inter-timestep and intra-timestep dependencies to be learned within Dynamic Bayesian Networks. We show that this combination of dependencies produces more accurate models than structures limited exclusively to inter-timestep or intra-timestep dependencies.

Structure learning has also been employed successfully to augment Bayes Net classifiers in ways that relax independence assumptions, and Bayesian Model Averaging has been employed to deal with structure uncertainty in small datasets. Friedman et al. (1997) augment the Naïve Bayes classifier with a tree-structure connecting the feature variables in addition to edges between the class and each feature. They call this new classifier Tree-Augmented Naïve Bayes (TAN). They find that the relaxation of the feature conditional independence assumption helps to improve the classification accuracy on numerous datasets. Selective Naïve Bayes (Langley and Sage 1994) relaxes the direct dependency requirement of each feature to the class variable, allowing built-in feature selection. Bayesian Model Averaging has also been employed for both TAN (Cerquides and de Mántaras 2003) and selective Naïve Bayes (Dash and Cooper 2004), resulting in increased accuracy for small datasets with large amounts of structure uncertainty.

The Selective Forest Augmented Naïve Bayes (SFAN) classifier provides further relaxation of the TAN constraints. It allows features to be interconnected by a forest structure and optionally dependent on the class variable. This class of structures is a simple case of Selectively Conditioned Forests. Heuristic approaches have been employed to create the same structure by pruning some of the edges from the TAN classifier (Sacha 1999), but our approach is the first to find the optimal MAP estimate or employ Bayesian Model Averaging.

## 3 Learning MAP SCF Structures

The focus of this work is the situation where a set of variables can be divided into two sets $\mathcal{S}_{t-1}$ and $\mathcal{S}_t$ (with values $\mathbf{S}_{t-1}$ and $\mathbf{S}_t$) and the variables of $\mathcal{S}_t$ are conditioned on the variables of $\mathcal{S}_{t-1}$. The graph structure, G, can similarly be split into the edges that are parents of variables in $\mathcal{S}_{t-1}$ and $\mathcal{S}_t$. We denote these two subgraphs as $G_{t-1}$ and $G_t$. Under the restriction that variables of $\mathcal{S}_{t-1}$ cannot have parents from $\mathcal{S}_t$, the choice of parents for variables in $\mathcal{S}_t$ and $\mathcal{S}_{t-1}$ are independent since no cycles can be formed spanning across both sets.

$$
\begin{aligned}
G^*_{\text{MAP}} &= \underset{G}{\operatorname{argmax}} \log P(\mathbf{S}_t|G_t,\mathbf{S}_{t-1}) + \log P(G_t) \\
&\quad + \log P(\mathbf{S}_{t-1}|G_{t-1}) + \log P(G_{t-1}) \\
&= \underset{G_t}{\operatorname{argmax}} \sum_{i:\,X_i\in\mathcal{S}_t} LS(\mathbf{X}_i|\mathbf{X}_{\pi_i},P_{\theta_i}) + \log P(\pi_i) \\
&\quad \cup \underset{G_{t-1}}{\operatorname{argmax}} \sum_{i:\,X_i\in\mathbf{S}_{t-1}} LS(\mathbf{X}_i|\mathbf{X}_{\pi_i},P_{\theta_i}) + \log P(\pi_i) \\
&= G^*_{t,\text{CMAP}|\mathbf{S}_{t-1}} \cup G^*_{t-1,\text{MAP}}
\end{aligned}
\tag{4}
$$

The MAP structure, $G^*_{\text{MAP}}$, can be obtained by combining the MAP structure for $G_{t-1}$ and the conditional MAP structure (CMAP) of $G_t$ given $\mathcal{S}_{t-1}$ as shown in Equation 4. The CMAP only differs from the MAP estimate in that variables of the CMAP can have parents in the disjoint condition set, $\mathcal{S}_{t-1}$, in addition to parents from within the same set, $\mathcal{S}_t$.

Selectively conditioned forests (SCFs) impose two additional constraints on the CMAP graph structure:

- Variable $X_{t,i}$ of $\mathcal{S}_t$ has at most $k$ parents from $\mathcal{S}_{t-1}$, which we denote $\pi_{t-1,i}$

- Variable $X_{t,i}$ has at most one parent, $\pi_{t,i}$, from $\mathcal{S}_t$ (i.e., a forest interconnect the variables of $\mathcal{S}_t$).

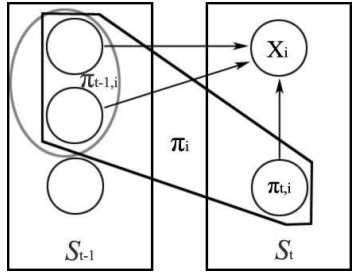

Figure 2: Sets of variables used in the algorithm.

Only intra-set parents, $\pi_t$, can create directed cycles. This simple observation allows efficient optimization. Given the intra-set parent, $\pi_{t,i}$, the inter-set parents, $\pi_{t-1,i}$, can be independently maximized by combinatorially evaluating the limited subset of parent variables in $\mathcal{S}_{t-1}$ (Equation 6).

$$\pi_t^* = \underset{\text{forest } \pi_t}{\operatorname{argmax}} \max_{\pi_{t-1}} \sum_i LS(\mathbf{X}_i|\mathbf{X}_{\pi_i}, P_\theta) + \log P(\pi_i)$$
$$= \underset{\text{forest } \pi_t}{\operatorname{argmax}} \sum_i \max_{\pi_{t-1,i}} LS(\mathbf{X}_i|\mathbf{X}_{\pi_{i,t}}, \mathbf{X}_{\pi_{t-1,i}}, P_\theta) + \log P(\pi_i)$$
$$= \underset{\text{forest } \pi_t}{\operatorname{argmax}} \sum_i LS(\mathbf{X}_i|\mathbf{X}_{\pi_{i,t}}, \mathbf{X}_{\pi_{t-1,i}^*(\pi_{t,i})}, P_{\theta_i}) + \log P(\pi_i) \quad (5)$$

$$\pi_{t-1,i}^*(\pi_{t,i}) = \underset{\pi_{t-1,i}}{\operatorname{argmax}} LS(\mathbf{X}_i|\mathbf{X}_{\pi_{t,i}}, \mathbf{X}_{\pi_{t-1,i}}, P_{\theta_i}) + \log P(\pi_i) \quad (6)$$

The dynamic algorithm follows from Equations 5 and 6. It presupposes each possible intra-set parent, $\pi_{t,i}$, and finds the best possible set of inter-set parents (Equation 6) for that parent $\pi_{t,i}$ (Step 1). We denote the combined best parent set of $X_i$ for inter-set parent $\pi_{t,i}$ as $\pi_i(\pi_{t,i})^*$. The score of these sets, $\lambda_i(\pi_{t,i})$ is used to weight the edge from the intra-set parent, $\pi_{t,i}$, to the child, $X_i$, in a directed graph (Step 2). The solution for Equation 5 is the maximum directed spanning forest (MDSF) of the graph[1]. The time-complexity of the algorithm is characterized by the $O(|\mathcal{S}_{t-1}|^k|\mathcal{S}_t|^2|D|)$ cost of performing LocalScore calculations.

---

[1] Each root in the forest also has a weight that contributes to the total structure score. We reduce to MDST with an additional vertex weight for the root considered in the maximization, using a modified version of the Chu-Liu-Edmonds algorithm (Chu and Liu 1965, Edmonds 1967).

---

**Algorithm 1** LearnSCF CMAP algorithm

**LearnSCF**($\mathcal{S}_t, \mathcal{S}_{t-1}$)

1. $\forall_{X_i \in \mathcal{S}_t, X_j \in \mathcal{S}_t \cup \{\emptyset\}}$

   $\pi_i(\pi_{t,i})^* \leftarrow \underset{\pi_i : \pi_i \cap \mathcal{S}_t \subseteq \{X_j\}}{\operatorname{argmax}} LS(\mathbf{X}_i|\mathbf{X}_{\pi_i}, P_{\theta_i}) + \log P(\pi_i)$

   $\lambda_i(\pi_{t,i})^* \leftarrow \underset{\pi_i : \pi_i \cap \mathcal{S}_t \subseteq \{X_j\}}{\max} LS(\mathbf{X}_i|\mathbf{X}_{\pi_i}, P_{\theta_i}) + \log P(\pi_i)$

   where also $|\pi_i \cap \mathcal{S}_{t-1}| \leq k$

2. Create full directed graph G=(V,E) where $V = \mathcal{S}_t$,

   $\forall_{i \neq j}$ weight($E_{j \rightarrow i}) \leftarrow \lambda_i(\pi_{t,j})^*$,

   $\forall_i$ weight($V_i) \leftarrow \lambda_i(\emptyset)^*$

3. Find the MDSF with weighted vertices for graph G yielding a *root* vertex and set of edges $E_{a_i \rightarrow b_i}$

4. $\pi_{\text{root}} = \pi_{\text{root}}(\emptyset)^*, \forall_{E_{a_i \rightarrow b_i} \in E_{MDSF}} \pi_{b_i} = \pi_{b_i}(a_i)^*$

---

## 4 SCF Bayesian Model Averaging

Using Bayesian Model Averaging, queries are probabilistically averaged over all possible models. For SCFs, Equation 3 can be rewritten in terms of inter-set and intra-set parents. Equation 7 follows using the same decomposition employed by Buntine (1991).

$$P(\tilde{\mathbf{X}}|\mathbf{X}) = \sum_{\pi_t} \sum_{\pi_{t-1}} \prod_i e^{LS(\tilde{\mathbf{X}}_i|\tilde{\mathbf{X}}_{\pi_i}, P_{\theta_i}|\mathbf{X}) + \log P(\pi_i)}$$

$$P(\tilde{\mathbf{X}}|\mathbf{X}) = \sum_{\pi_t} \prod_i \sum_{\pi_{t-1}} e^{LS(\tilde{\mathbf{X}}_i|\tilde{\mathbf{X}}_{\pi_i}, P_{\theta_i}|\mathbf{X}) + \log P(\pi_i)} \quad (7)$$

The inner summation of Equation 7 averages over all possible inter-set parents of a variable given an intra-set parent. Again, this is possible because inter-set parents can create no directed cycles. We can now employ the Matrix Tree Theorem for directed graphs to efficiently average over all tree (or forest) structures.

**Theorem 1 (Tutte (1948))** *If we construct a matrix A such that*

$$a_{ij} = \begin{cases} \sum_k w_{k,j} & \text{if } i = j \\ -w_{i,j} & \text{if } i \neq j \end{cases}$$

*and if $A_{-k}$ is the matrix created by removing row $k$ and column $k$ of $A$ then:*

$$\det(A_{-k}) = \sum_{T_n} \prod_{(i,j):E(i \rightarrow j) \in T_k} w_{i,j} \quad (8)$$

*where $T_n$ is each directed spanning tree rooted at $k$.*

The efficient $O(|\mathcal{V}|^3)$ computation obtained from Theorem 1 can be generalized to forests with weighted roots by augmenting the graph with a new root $X_0$ and edges $w_{0,j} = \text{rootweight}_j$.

Equation 7 can then be efficiently calculated[2] using

---

[2] In many structures the choice of direction for a single edge is arbitrary. This can lead to an ill-conditioned matrix. A heuristic (e.g., the edge direction in the MAP estimate) can be employed to direct such edges.

the following weight settings.

$$w_{i,j} = \sum_{\pi_{\mathbf{j}} : \pi_{\mathbf{j},t} = X_i} e^{LS(\tilde{\mathbf{X}}_j | \tilde{\mathbf{X}}_{\pi_{\mathbf{j}}}, P_\theta | \mathbf{X}) + P(\pi_{\mathbf{j}})} \qquad (9)$$

$$w_{0,j} = \sum_{\pi_{\mathbf{j}} : \pi_{\mathbf{j},t} = \emptyset} e^{LS(\tilde{\mathbf{X}}_j | \tilde{\mathbf{X}}_{\pi_{\mathbf{j}}}, P_\theta | \mathbf{X}) + P(\pi_{\mathbf{j}})} \qquad (10)$$

where $|\pi_{\mathbf{j},\mathbf{t-1}}| \leq k$.

## 5 SCF Dynamic Bayesian Networks

Our selectively conditioned forest algorithms were inspired by Dynamic Bayesian Networks. The class of graphs for the repeating slice of the DBN is a mixture of inter-timestep dependencies connecting variables in $X_{t-1}$ to variables in $X_t$ and intra-timestep dependencies connecting variables within $X_t$. This formulation lends itself naturally to SCFs where $X_{t-1}$ is the conditional set and $X_t$ is the target set. SCFs allow an intra-timestep forest structure and a limited number of inter-timestep parents. An example from the class of SCF DBNs is shown in Figure 3.

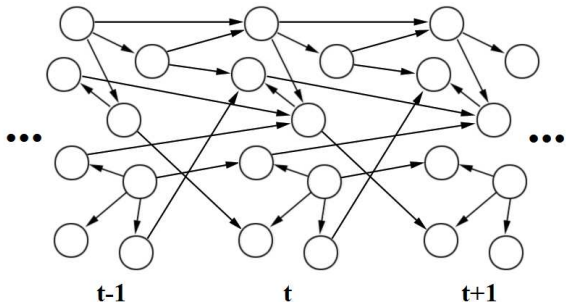

Figure 3: A partially unrolled SCF DBN with an intra-timestep forest structure and inter-timestep parents limited to two.

For a DBN with $m$ variables in each timestep and inter-timestep dependencies limited to k, the running times of MAP SCF and BMA SCF are both $O(m^{k+2}|D|)$. The resulting structure (or average over structures) optimizes $P(\mathbf{X_{t+1}}|\mathbf{X_t}, G)P(G)$, making SCFs well-suited for prediction.

### 5.1 Experiments

We compare average withheld predictive performance, $\overline{\log P(\mathbf{X_t}|\mathbf{X_{t-1}})}$, of MAP SCF and BMA SCF with MAP estimates of subclasses of SCF on two different temporal datasets. The first (Office) has six variables that are discretizations of sensor values measuring keyboard and mouse usage, sound levels, motion, and the status of the office door at minute-level granularity. There are approximately 80,000 timesteps of data. We use the SCF algorithm and train over 99% of the data and test on the remaining withheld data. The second dataset is the Bayesian Automated Taxi (BAT) DBN (Forbes et al. 1995), a synthetic DBN modeling vehicle status that has 28 variables. We generate 500 and 1000 timesteps of training data and 1000 timesteps of testing data. We train using BDeu with sample size 20 and a uniform structure prior.

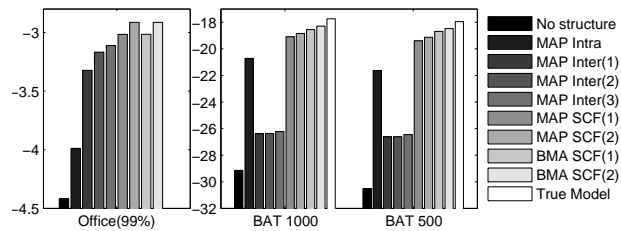

Figure 4: Average withheld log-probability of next timestep data given current timestep data.

The results are shown in Figure 4. The *No structure* model uses the marginal probabilities for each variable. The *Intra* model learns a forest structure to connect variables within the same timestep. The *Inter(k)* model allows each variable at most k parents from the previous timestep.

Of the structure MAP estimates, MAP SCF performs the best of these classes, which we would expect since it is learning a superclass of the other models' structures. Of perhaps more significance, in both datasets the *MAP SCF(1)* model outperforms all other *Inter* models, suggesting the importance of intra-temporal dependencies in DBNs. BMA SCF is equivalent to MAP SCF in the Office dataset. However, for the smaller BAT experiments where structure uncertainty is high, BMA SCF outperforms MAP SCF significantly.

## 6 Selective Forest-Augmented Naïve Bayes

We now show how SCFs can be applied to Bayesian Network classification to create a generalization of the TAN classifier that performs similarly on "clean" data with only relevant features, while offering built-in feature selection for improved performance on data with random, irrelevant features. The resulting classifiers, selective tree-augmented Naïve Bayes (STAN) and selective forest-augmented Naïve Bayes (SFAN), are not new, but we provide the first method for BMA of SFAN and the first method and evaluation of exact MAP estimated SFAN and STAN.

Figure 5 illustrates the characteristic differences between Naïve Bayes and augmented Naïve Bayes models. Selectivity integrates feature selection into the structure learning of augmented Naïve Bayes classi-fiers. MAP SFAN can be learned using the MAP SCF algorithm with the class as the condition set, $\mathcal{S}_{t-1}$, and the features as the target set, $\mathcal{S}_t$. MAP STAN and Forest-Augmented Naïve Bayes (FAN) can also be learned using simple modifications to the MAP SCF algorithm that impose different structure restrictions.

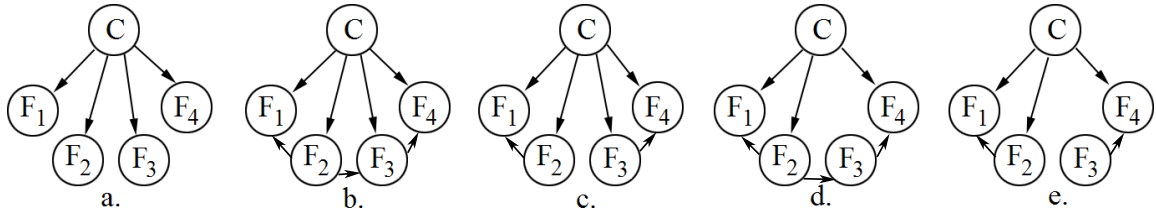

Figure 5: Naïve Bayes classifiers with augmentation. (a) Naïve Bayes (b) tree-augmented NB (TAN) (c) forest-augmented NB (FAN) (d) selective tree-augmented NB (STAN) (e) selective forest-augmented NB (SFAN).

Though SFAN is a generalization of the TAN classifier, increased generality in augmented Bayes classifiers can lead to decreased classification accuracy (Friedman et al. 1997). This is because generative models maximize the joint likelihood while the Bayes optimal classifier maximizes conditional likelihood of the class variable (Y) given the feature values (X).

$$P(\mathbf{X}, \mathbf{Y}|G) = P(\mathbf{Y}|\mathbf{X}, G) + P(\mathbf{X}|G) \quad (11)$$

MAP selective-augmented Bayes classifiers risk yielding very small Markov blankets for the class variable by maximizing the joint likelihood primarily through the $P(\mathbf{X}|G)$ term. In other words, the features explain away most of the other features without needing the class variable. Friedman et al. (1997) require the class variable to be a parent of each feature to deal with this problem, but this can degrade classification accuracy in the presence of irrelevant features. While in our experiments SFAN doesn't underperform against FAN, we propose addressing this issue by adjusting the graph structure prior to prefer including the class variable as a parent without requiring it. We can modify the objective function to penalize for excluding the class variable.

$$\begin{aligned}
G^* &= \operatorname*{argmax}_G \log \int_\Theta P(G|\mathbf{D}) \, d\Theta \\
&= \operatorname*{argmax}_G \log \int_\Theta P(\mathbf{D}|G, \Theta) P(G, \Theta) \, d\Theta \\
&= \operatorname*{argmax}_G \left[ \sum_{i \in \mathcal{S}_t} LS(\mathbf{X_i}|\mathbf{X}_{\pi_i}, P_\theta) + P(\pi_i) + \text{penalty}(\pi_i) \right]
\end{aligned} \quad (12)$$

Using an indicator function, $\delta(.)$, which is one when the statement is true and zero otherwise, we can parameterize the penalty by $\alpha$.

$$\text{penalty}(\pi_i) = -\alpha \delta(\mathcal{S}_{t-1} \notin \pi_i) \quad (13)$$

Setting $\alpha = \infty$ yields the FAN classifier (Figure 5c), while setting $\alpha = 0$ yields the SFAN classifier (5e).

In sparse datasets where some features are independent of the class variable and other features are useful discriminatively, but only weakly so, we would expect a value of $\alpha$ between these two extremes to produce better results than either the FAN or SFAN classifier.

### 6.1   Experiments

We evaluate our classifiers using 8 UCI datasets (Newman et al. 1998). We remove any examples with missing features. We perform 10-fold cross-validation and entropy-based discretization of continuous variables (Fayyad and Irani 1993) for each fold. We use the BDeu prior within the MAP SCF algorithm and use a prior weight of 10 and a uniform structure prior.

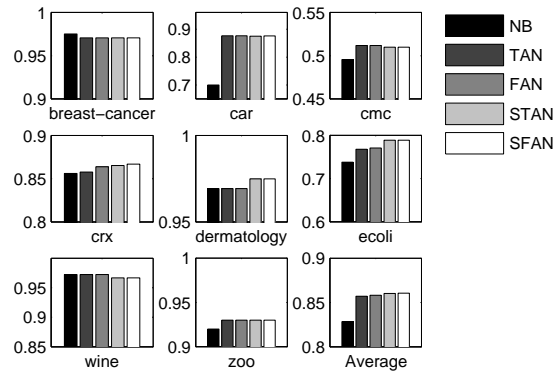

Figure 6: Classification accuracy Rates of Naïve Bayes and augmented Naïve Bayes classifiers on UCI datasets

The results of our first experiment in Figure 6 show the classification accuracy rates of the various Naïve Bayes-based classifiers on eight different UCI datasets and their average classification accuracy. We find a significant difference in average classification accuracy between the Naïve Bayes classifier and each of the augmented classifiers, but insignificant differences between augmented classifiers.

The results of our second set of experiments in Figure 7 and Figure 8 show the robustness of the differ-ent classifiers to irrelevant features. Under the same settings as the first experiment, we add 5 (Figure 7) and 20 (Figure 8) uniform random binary noise features. NB, TAN, and FAN all have significant accuracy degradations as the amount of irrelevant features increases, while the STAN and SFAN models effectively ignore the irrelevant features (Figure 9). Other methods exist for performing feature selection (e.g., wrapper-based (Kohavi and John 1997)) that often increase the time complexity of the algorithm and use heuristics, but MAP SCFs incorporate it into part of the structure learning process for structure-augmented classifiers. This addition of selectivity does not increase the run-time complexity versus TAN (without feature selection), and employs statistically justified Bayesian scoring metrics.

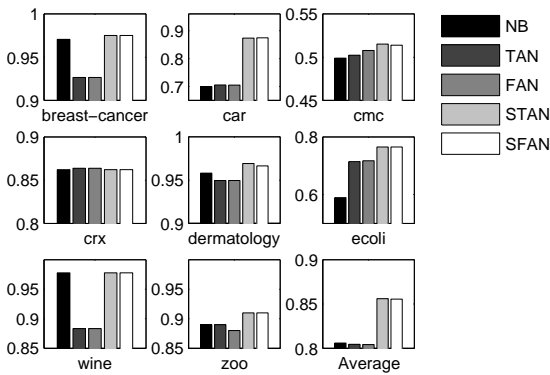

Figure 7: Classification accuracy rates on UCI datasets with 5 irrelevant features added.

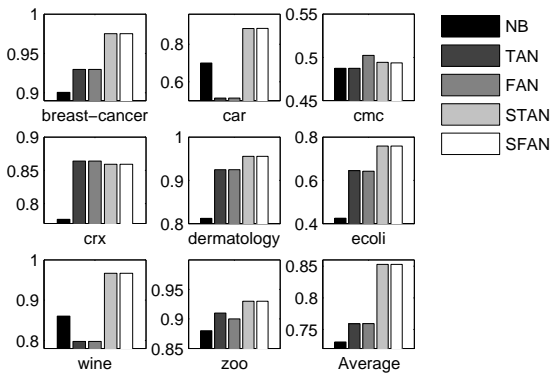

Figure 8: Classification accuracy rates on UCI datasets with 20 irrelevant features added.

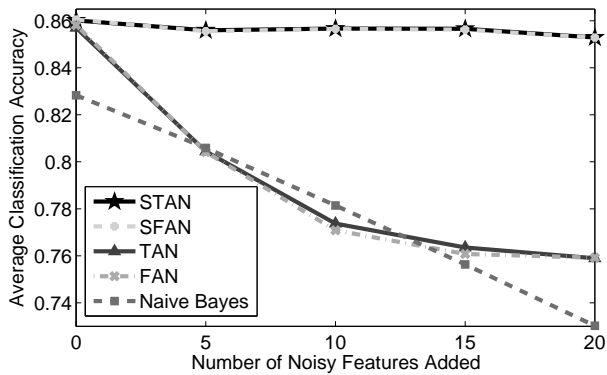

Figure 9: Average classification accuracy rates on UCI datasets versus number of irrelevant features added.

Our final experiment demonstrates a situation where a selective classifier (SFAN) can underperform against its non-selective counterpart (FAN), but by employing the exclusion penalty (Equation 13) better classifica-tion accuracy than either the original SFAN or FAN classifier can be achieved. We construct a synthetic dataset with 10 binary features that are weakly dependent on the class variable ($p(F_i = C) = .6$), and 20 random binary features that are independent of the class variable. We generate 100 examples to train an SFAN classifier under varying exclusion penalty constants and test on a set of separately generated 100 examples. We average over 100 experiments.

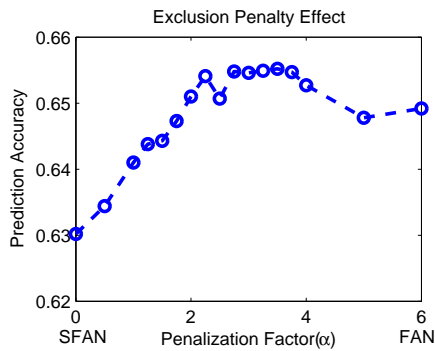

Figure 10: The exclusion penalty versus classification accuracy for learning features weakly correlated to the class among additional irrelevant features

The results are shown in Figure 10. At the extremes of the graph we have the unpenalized SFAN classifier ($\alpha = 0$) and the FAN classifier ($\alpha = 6$). Both FAN and SFAN underperform against the SFAN classifier with exclusion penalty $\alpha \in [2, 4]$. We believe some real datasets will have similar "happy mediums" where the discriminative benefit of including more potentially relevant features is balanced against the detrimental effect of including too many irrelevant features in the class variable's Markov blanket .

We've shown that the built-in feature selection of SFAN provides equivalent classification accuracy to FAN in relatively clean data and far greater robustness to noisy, irrelevant features. Additionally, we've described how the graph prior can be modified to better balance the benefits of a large Markov blanket for the class variable versus the detriments of including many irrelevant features. We believe this evidence and the similar run-time and implementation complexities suggest SFAN should be preferred to the commonly employed TAN classifier for Bayes Net classification.

# 7    Conclusions and Future Work

We have shown how two existing classes of structures that allow efficient structure learning, tree structures and limited fixed-orderings, can be combined while still allowing efficient structure learning. We call this combined class of structures *selectively conditioned forests*. We have presented algorithms for efficiently learning MAP estimates of SCFs and Bayesian Model Averaging for the class of SCFs. Prior to this work, SCFs could only be learned approximately.

We demonstrated the usefulness of this class of structures in two domains. Applied to Dynamic Bayesian Networks, SCFs learn structures with both intra-timestep and inter-timestep dependencies. We showed empirically that this combination of dependencies improves predictive model accuracy. We additionally showed that Bayesian Model Averaging helps to further improve predictive model accuracy for small training datasets. Applied to generative classification models, SCFs yield Selective Forest-Augmented Naïve Bayes classifiers. We showed empirically that these classifiers perform as well as TAN on noise-free data, and better than TAN on noisy data. We believe that this noise robustness, along with the comparable run-time costs and implementation complexity, make SFAN a preferable classifier to TAN.

Our plans for the continuations of this work are twofold. First, we would like to apply SCFs to datasets with missing values so that our learning algorithms can be extended to Hidden Markov Models for instance. Second, further empirical analysis of the SFAN classifier is warranted. Specifically, we would like to empirically measure the benefits of the exclusion penalty prior and Bayesian Model Averaging over SFANs for the important problem of classification.